\documentclass[journal]{IEEEtran}
\usepackage{amsmath}
\usepackage{cite}
\usepackage{graphicx}
\usepackage{tabularx}
\usepackage{booktabs}
\usepackage{caption}
\usepackage{tikz}
\usetikzlibrary{arrows}
\newcolumntype{C}{>{\centering\arraybackslash}X}
\begin{document}

\title{Performance of Three Slim Variants of \\
	The Long Short-Term Memory (LSTM) Layer}
       \author{\IEEEauthorblockN{Daniel~Kent and Fathi~Salem}\\
	       \IEEEauthorblockA{Wireless and Video Communications (WAVES) Lab\\
	Circuits, Systems, and Neural Networks (CSANN) Lab\\
        Department of Electrical and Computer Engineering\\
	Michigan~State~University\\
	East~Lansing,~Michigan,~United States of America}}

\maketitle

\begin{abstract}
	The Long Short-Term Memory (LSTM) layer is an important advancement
in the field of neural networks and machine learning, allowing for effective
training and impressive inference performance. LSTM-based neural networks
have been successfully employed in various applications such as speech processing and language translation.
The LSTM layer can be simplified by removing certain components, potentially
speeding up training and runtime with limited change in performance. In particular, the recently introduced variants, called  
SLIM LSTMs, have shown success in initial experiments to support this view. 
Here, we perform computational analysis of the validation accuracy of a convolutional plus
recurrent neural network architecture using comparatively the \textit{standard} LSTM and three 
SLIM LSTM layers.
We have found that some realizations of the SLIM LSTM layers can potentially perform
as well as the \textit{ standard} LSTM layer for our considered architecture.
\end{abstract}

\section{Introduction}
\subsection{LSTM Architecture Overview}
The Long-Short Term Memory (LSTM) layer is a type of Recurrent Neural Network
(RNN) first proposed by Hochreiter and Schmidhuber in 1997
\cite{NeuralComputation:001}. More recent formalisms and explorations of LSTM RNN are described in 
\cite {Odyssey2016} and the refererencs therein. Successful example applications include speech processing, e.g.,  
\cite {boulanger2012modeling} and \cite {GoogleResearch:001}
and language translation, e.g.,  \cite{DBLP:journals/corr/JohnsonSLKWCTVW16}.

The standard LSTM layer has three gates:
an input gate $i_t$ , a forget gate $f_t$ , and an output 
gate $o_t$. Each gate is a replica of the ``input Block" RNN. The overall equations of this standard LSTM layer
are described in \cite {Odyssey2016}, and the references therein. Here, we follow the presentation in \cite{SalemMemo:001, Salem2018:001}, 
where one splits the 3 gating equations from the memory cell and the``input block" equations for suitability of the development in the next sections. \\
The 3 gating equations are:
\begin{flalign}
i_t &= \sigma_{in}(W_i x_t + U_i h_{t-1} + b_i) \\
f_t &= \sigma_{in}(W_f x_t + U_f h_{t-1} + b_f) \\
o_t &= \sigma_{in}(W_o x_t + U_o h_{t-1} + b_o)
\end{flalign}
and the cell-memory/input block equations are:
\begin{flalign}
\tilde{c_t} &= \sigma(W_c x_t + U_c h_{t-1} + b_c) \\
c_t &=f_t \odot c_{t-1} + i_t \odot \tilde{c_t} \\
h_t &= o_t \odot \sigma(c_t) 
\end{flalign}
where equations (1-3) are the gating singals, eqaution (4) is the ``Input Block" equation, equation (5) is the memory-cell equation, and equation (6) is the hidden unit/activation equation. It is noted that the gate equations, each is a replica of the Input Block eqaution (4). 
In this notation, $x_t$ is the input vector (sequence), say of dimension $m$, 
the memory "state" $c_t$ is of dimension $n$, as are the three gate signal vectors $i_t$, $f_t$, and $o_t$, and also he hidden unit/activation $h_t$. Thus, the sizes of the parameters: matrices $W_*$, $W_*$, and bias vectors, $b_*$ are easily specified. The set of equations (1-6) constitute the definition of the (\textit{standard}) LSTM layer considered here. In the next section, we shall focus on simplified gating of equations (1-3) to define the three LSTM variants of interest here. 

\subsection{SLIM LSTM Variants Overview}
More recently, a host of new variants with aggressive reduction of parameters of the LSTM layer have shown reasonable initial success, see \cite{salem2016reduced, LuSalem2017, HeckSalem2017, DeySalem2017, SalemMemo:001}. These mosaic of variants are referred to as SLIM LSTMs \cite{Salem2018:001}.

Here, we explore further the first three SLIM LSTM variants, denoted as LSTM1,
LSTM2, and LSTM3 as termed in \cite{salem2016reduced, LuSalem2017, SalemMemo:001, Salem2018:001}.
\hfil \break
\subsubsection{Slim LSTM\_1}
(or simply LSTM1) removes the input signal and corresponding weights
from the gating signals in the layer as per the following parameter-reduced gating equations:

\begin{flalign*}
i_t &= \sigma_{in}(U_i h_{t-1} + b_i)\\
f_t &= \sigma_{in}(U_f h_{t-1} + b_f)\\
o_t &= \sigma_{in}(U_o h_{t-1} + b_o)
\end{flalign*}
These gating equations replace equations (1-3) to generate the LSTM1 layer.
\hfil \break
\subsubsection{Slim LSTM\_2}

(or simply LSTM2) removes both the bias and input signals and their
corresponding weights as per the following reduced equations:

\begin{flalign*}
i_t &= \sigma_{in}(U_i h_{t-1}) \\
f_t &= \sigma_{in}(U_f h_{t-1}) \\
o_t &= \sigma_{in}(U_o h_{t-1}) \\
\end{flalign*}
These gating equations replace equations (1-3) to generate the LSTM2 layer.
\hfil \break
\subsubsection{Slim LSTM\_3}

(or simply LSTM3) removes both the input signal and the hidden unit and their
corresponding weights as per the following reduced equations:

\begin{flalign*}
i_t &= \sigma_{in}(b_i) \\
f_t &= \sigma_{in}(b_f) \\
o_t &= \sigma_{in}(b_o) 
\end{flalign*}
These gating equations replace equations (1-3) to generate the LSTM3 layer.

\section{Experiment Parameters}
\subsection{Neural Network Parameters}
\begin{figure}[ht]
\begin{center}
    \includegraphics[height=5cm]{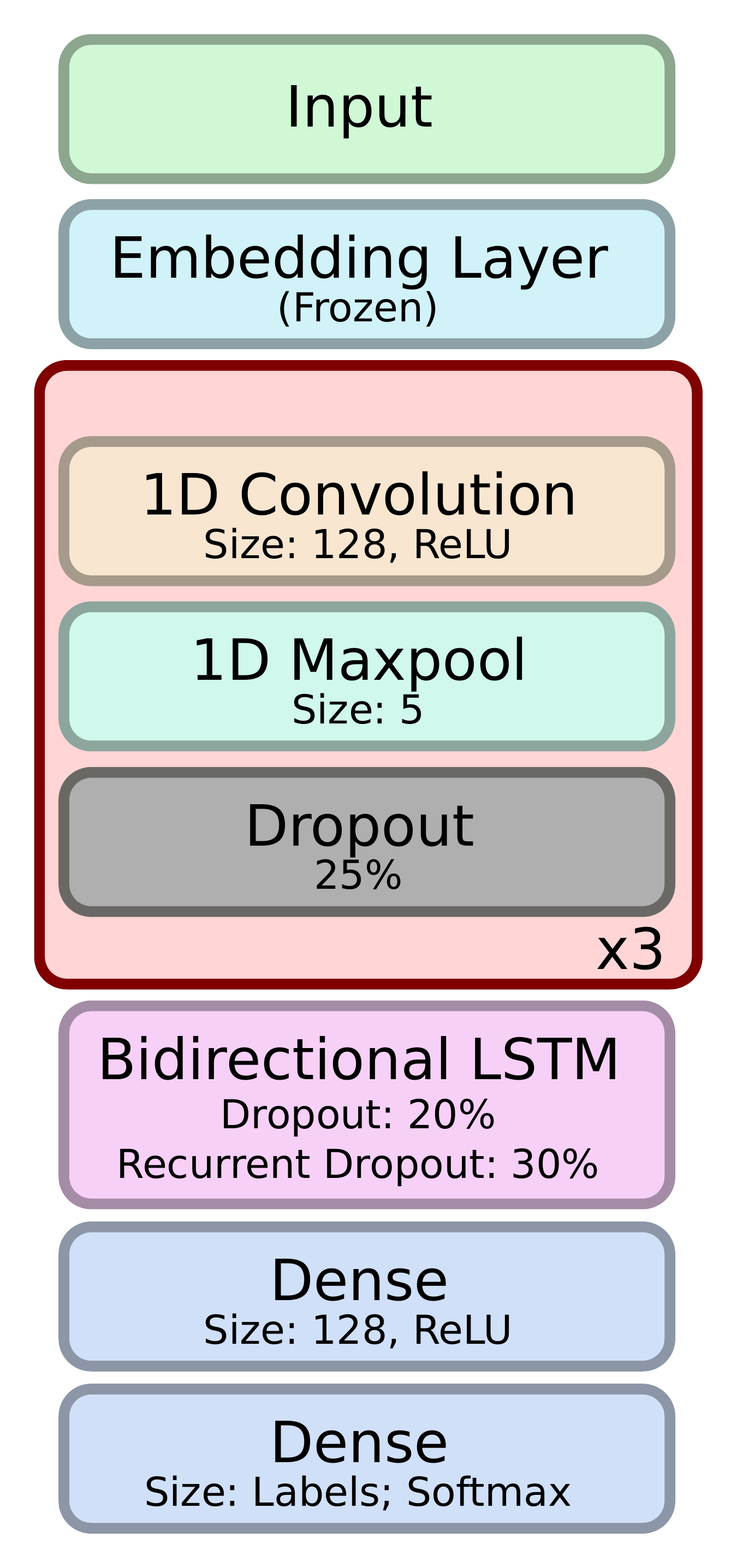}
    \caption{Neural Network Architecture Diagram}
    \label{fig:neuralnet}
\end{center}
\end{figure}

The Neural Network Architecture used in this work is depicted in Fig. \ref{fig:neuralnet}. It is a hybrid convolutional plus
bidirectional recurrent neural network. There is an input layer, followed
by an Embedding layer that is pre-trained on the GloVe dataset \cite{Dataset:GloVe}
that is then followed by three sets of one-dimensional convolutional and
maxpooling layers with dropout, followed by a Bidirectional LSTM layer with 20\%\
dropout and 30\% recurrent dropout, followed
by two densely connected layers.
The assembled Neural Network architecture was trained on the 20-Newsgroup dataset \cite{Dataset:Newsgroup}.
\subsection{Hardware and Software}
The neural network architecture was built using Keras 2.0 running in Python 2.7.14 on 
a workstation running Ubuntu 17.10 x86\_64, using code based off of the Keras
Pretrained Word Embeddings example code, with modifications made to accomodate
the additional features needed to test the parameters outlined in this paper.
\subsection{Tested Parameters}
We have tested three types of ``variables": (i)  the LSTM variants, (ii) the activation function for the 
Bidirectional LSTM layer variant, and (iii) the learning rate. LSTM variants tested were the base (LSTM), LSTM1, LSTM2, and LSTM3. Activation
functions tested were the hyperbolic tangent function (tanh), Linear activation,
Sigmoid activation, ReLU, and Softmax. Learning rates test were 2e-3, 1e-3, and
5e-4. .

\section{Results and Discussion}

\subsection{Validation Accuracy Results}
\begin{table*}
     \caption{Validation Accuracy After 100 Epochs}
     \label{my-labelI}
\begin{tabularx}{\textwidth}{@{}l*{5}{C}c@{}} 
    \toprule
    \textbf{LSTM Activation}  & \textbf{Learning Rate}  & \textbf{LSTM} & \textbf{LSTM1} & \textbf{LSTM2} & \textbf{LSTM3}  \\   \midrule
    \textbf{Tanh} & 2.00E-003 & 73.793\% & 73.718\% & 72.318\% & 74.469\%  \\
    \textbf{Tanh} & 1.00E-003 & 72.443\% & 72.668\% & 72.618\% & 71.768\%  \\
    \textbf{Tanh} & 5.00E-004 & 73.518\% & 72.343\% & 71.443\% & 70.643\%  \\
    \textbf{Linear} & 2.00E-003 & 4.501\% & 4.376\% & 4.776\% & 72.668\%  \\
    \textbf{Linear} & 1.00E-003 & 72.543\% & 72.468\% & 69.742\% & 73.218\%  \\
    \textbf{Linear} & 5.00E-004 & 72.493\% & 71.218\% & 71.993\% & 70.893\%  \\
    \textbf{Sigmoid} & 2.00E-003 & 73.093\% & 72.343\% & 73.243\% & 71.818\%  \\
    \textbf{Sigmoid} & 1.00E-003 & 71.118\% & 70.943\% & 71.618\% & 70.993\%  \\
    \textbf{Sigmoid} & 5.00E-004 & 70.968\% & 69.792\% & 69.967\% & 68.392\%  \\
    \textbf{Softmax} & 2.00E-003 & 70.393\% & 60.965\% & 4.501\% & 26.132\%  \\
    \textbf{Softmax} & 1.00E-003 & 69.717\% & 66.317\% & 47.787\% & 58.690\%  \\
    \textbf{Softmax} & 5.00E-004 & 63.941\% & 49.862\% & 29.482\% & 48.012\%  \\
    \textbf{ReLU} & 2.00E-003 & 68.367\% & 68.467\% & 4.376\% & 73.043\%  \\
    \textbf{ReLU} & 1.00E-003 & 73.143\% & 73.618\% & 72.118\% & 71.943\%  \\
    \textbf{ReLU} & 5.00E-004 & 71.443\% & 72.593\% & 72.468\% & 73.118\%   \\    \bottomrule
\end{tabularx}
\end{table*}

\begin{table*}
     \caption{Validation Loss After 100 Epochs}
     \label{my-labelII}
     \begin{tabularx}{\textwidth}{@{}l*{5}{C}c@{}}
    \toprule
    \textbf{LSTM Activation}  & \textbf{Learning Rate}  & \textbf{LSTM} & \textbf{LSTM1} & \textbf{LSTM2} & \textbf{LSTM3}  \\  \midrule
    \textbf{Tanh} & 2.00E-003 & 1.26626372355 & 1.17280639938 & 1.30214396743 & 1.19286979217  \\
    \textbf{Tanh} & 1.00E-003 & 1.2533884461 & 1.24007106198 & 1.25131325291 & 1.28025298847  \\
    \textbf{Tanh} & 5.00E-004 & 1.12991302266 & 1.25897071954 & 1.25066449667 & 1.25003132021  \\
    \textbf{Linear} & 2.00E-003 & 2.99683079114 & 2.9973659225 & 2.99631883473 & 1.25062232564  \\
    \textbf{Linear} & 1.00E-003 & 1.12306529774 & 1.17703753339 & 1.22609648397 & 1.1734144355  \\
    \textbf{Linear} & 5.00E-004 & 1.11253687446 & 1.34505273065 & 1.14925828157 & 1.29449143705  \\
    \textbf{Sigmoid} & 2.00E-003 & 1.17115093154 & 1.18445872471 & 1.18269565109 & 1.23696543557  \\
    \textbf{Sigmoid} & 1.00E-003 & 1.29201206186 & 1.30768913518 & 1.27823424885 & 1.2994762417  \\
    \textbf{Sigmoid} & 5.00E-004 & 1.25363427584 & 1.29810960694 & 1.20362862963 & 1.34441118063  \\
    \textbf{Softmax} & 2.00E-003 & 1.17432751731 & 1.26089545492 & 2.99665749279 & 2.14994023913  \\
    \textbf{Softmax} & 1.00E-003 & 1.2308979069 & 1.30809664535 & 1.65336642154 & 1.38786871599  \\
    \textbf{Softmax} & 5.00E-004 & 1.29078156044 & 1.61376436578 & 1.91619378461 & 1.62440377285  \\
    \textbf{ReLU} & 2.00E-003 & 1.22890987945 & 1.30157855895 & 2.99658317911 & 1.11470258686  \\
    \textbf{ReLU} & 1.00E-003 & 1.10078433252 & 1.1446512137 & 1.0409039263 & 1.32684082522  \\
    \textbf{ReLU} & 5.00E-004 & 1.27224681913 & 1.14206915571 & 1.10311798523 & 1.20242762065  \\   \bottomrule
\end{tabularx}
\end{table*}
The conducted experiments are summarized in Table I.  Based on all the cases, LSTM3 with a hyperbolic tangent activation performed the best in terms of maximum achieved accuracy, both when considering only the data for the 100th epoch, as well as across all epochs.

Based on the average validation accuracy for all learning rates and all LSTM variants, it appears as though the Hyperbolic Tangent activation (tanh) generally works the best across all LSTM variants, though ReLU and Linear activations worked when the learning rate was 1e-3 or 5e-4, and Sigmoid activations worked reasonably well under all the learning rates selected. The softmax activation did not appear to work well at all, only achieving a validation accuracy over 50\% at any training point in 7 of 12 cases, and only achieving a 100-epoch validation accuracy over 50\% in only 6 of 12 cases.

For all LSTM variants, the standard LSTM model appears to work the best across all the tested activations and learning rates based on the average 100 epoch validation accuracy and the average maximum achieved validation accuracy. However, LSTM1 and LSTM3 on average performed only slightly worse than standard LSTM. LSTM2 appeared to perform the worst on average, only achieving a maximum validation accuracy of about 59.23\% on average, and 53.90\% after 100 epochs.

\begin{figure}[ht]
    \includegraphics[width=\linewidth]{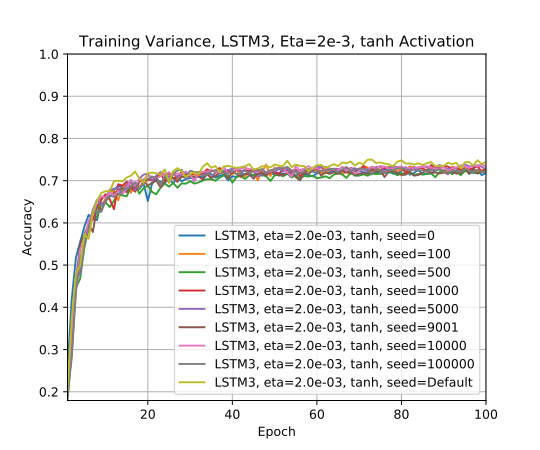}
    \caption{LSTM3 Training Variance, Tanh Activation, eta=2e-3}
    \label{fig:variance}
\end{figure}

To determine how much the results for the best case (LSTM3, Hyperbolic Tangent, learning rate 2e-3) vary based on initial seed, we re-ran the same case ten times with ten different initial seeds: 0, 100, 500, 1000, 5000, 9001, 10000, and 100000. Note that due to some non-deterministic behavior in (the backend) Tensorflow due to the use of CuDNN, some of the variance cannot be controlled by setting a random seed. Additionally, the code uses a time-based seed as a default for training if a seed isn’t specified (denoted as “Default” in Fig. \ref{fig:variance}).

As depicted in Fig. \ref{fig:variance}, the variance was high enough to suggest that the less than 1\% performance margin
enjoyed by the best LSTM3 case over the best LSTM case is not enough to conclusively assert that LSTM3 is superior
to LSTM for our architecture.

\subsection{Validation Loss Results}

While accuracy is one way to measure training effectiveness, loss is another important metric to judge a specific neural network performance. It is important to note that the loss expressions are only relative as they invlove different size network parameters!
In this case, it turns out that the minimum loss was achieved not by a hyperbolic tangent activation or an LSTM3 variant, but rather by a standard LSTM with linear activation. At 100 epochs, LSTM2 with ReLU activation appears to achieve the lowest loss. While ReLU activation performed well (but not the best). Like with validation loss, the performance figures for softmax activation are not promising, with a minimum average loss of 1.6 versus the next closest of 1.15 for linear activation. 
\subsection{Breakdowns with Linear Activation and High Eta}

\begin{figure}[ht]
    \includegraphics[width=\linewidth]{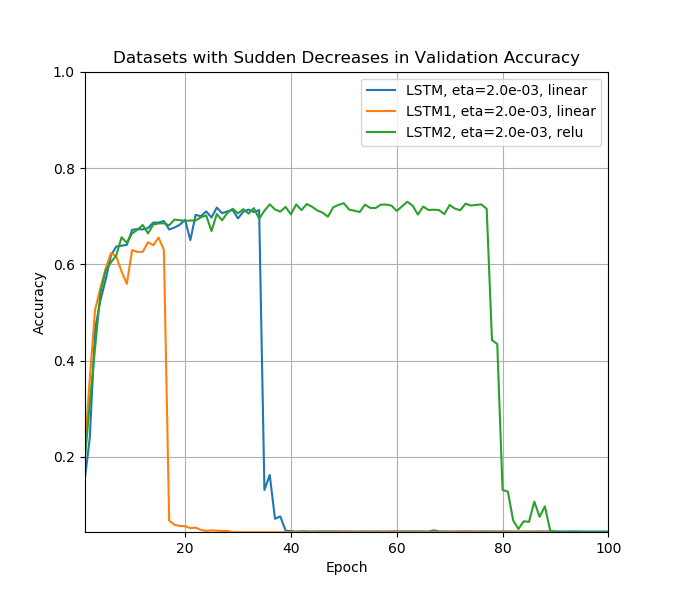}
    \caption{LSTM Training, Linear Activation, eta=2e-3}
    \label{fig:failure1}
\end{figure}

\begin{figure}[ht]
    \includegraphics[width=\linewidth]{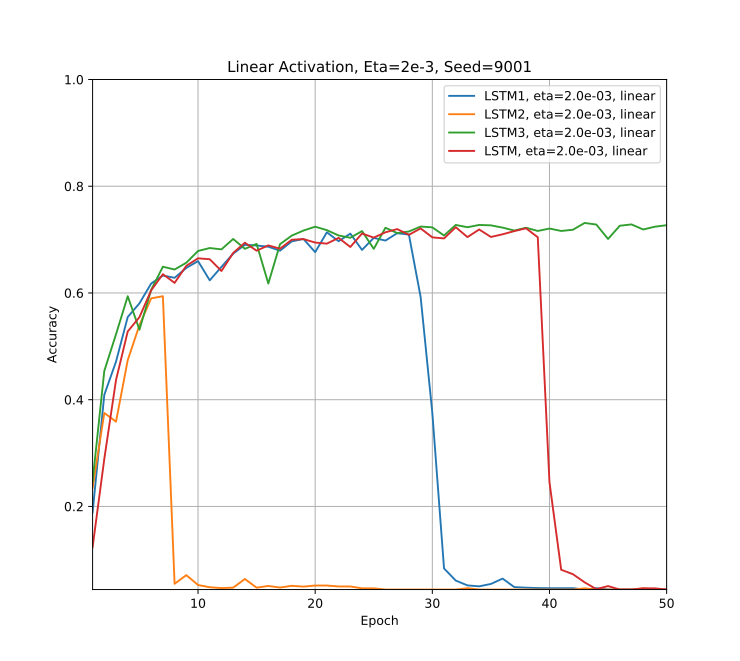}
    \caption{LSTM Training, Linear Activation, eta=2e-3}
    \label{fig:failure2}
\end{figure}
During training, it was observed that many of the test cases with linear
activations and a high (2e-3) learning rate would train normally up until a
certain epoch, at which point the validation accuracy would reduce to less than
5\% and no longer improve, and the loss would remain high.
To verify that this was not an issue with the particular starting seed for the
results in Fig. \ref{fig:failure1}, we re-ran the case with a different random seed;
these results, shown in Fig. \ref{fig:failure2}, indicate that this is not
a transient issue that only occurs in some cases.

\section{Conclusions}
\subsection{LSTM Variant Performance}
Based on the comparison of average validation accuracy across learning rates and
activation functions, LSTM3 appears to have the best average accuracy out of
all of the reduced LSTM variants, and additionally does not appear to vary
significantly from the base LSTM layer's performance. While some tests indicate
that LSTM3 was the best variant overall, training variance was high enough that
the results merely suggest that LSTM3 isn't strictly better than the base LSTM
in terms of validation performance and loss. Still, if this quality holds
up in other architectures, it could provide a basis for using LSTM3 by default
in performance-critical roles.
\subsection{Activation Function Performance}
Based on the comparison of average validation accuracy across learning rates
and LSTM layer types, the hyperbolic tangent function appears to have the
best average accuracy compared to all the other tested activation functions.
If this quality holds true for other architectures, it may provide justification
for using hyperbolic tangent by default instead of a linear activation as some
frameworks do.
\subsection{Training Breakdowns}
Based on the breakdown of validation accuracy in certain cases after a number
of epochs, it is reasonable to assume that a certain number of epochs of
training should be completed on any model before drawing conclusions on its
training behavior. This issue relates to issues of robustness of the generated neural model and their potential failures. 

\subsection{Rationale for the LSTM3 Strong Performance}
The LSTM3 layer does not \textit{apriori} impose structural form on the gating signal. By using only the biases in the gates, the learning technique (in this case, it is the Backpropagation Through Time (BPTT) \cite{Odyssey2016}), has potentially more freedom to steer the adaptng biases towards achieving a (relatively) lower loss. The adaptive process for the parameters will invlove the input signal profile, hidden units. In contrast, the standard LSTM makes the imposition of a definite structure that may not be convenient in all experiments or datasets, see \cite {Salem2018:001}. Of course, the choice of the ``optimal" hyper-parameters in each LSTM variant has the potential of achieving strong performance in each variant. 

\section*{Acknowledgement}

This work was supported in part by the National Science Foundation under grant No. ECCS-1549517.

\end{document}